# Machine Learning Techniques for Pattern Recognition in High-Dimensional Data Mining

Pochun Li
Northeastern University
Boston, USA

*Abstract*—This paper proposes a frequent pattern data mining algorithm based on support vector machine (SVM), aiming to solve the performance bottleneck of traditional frequent pattern mining algorithms in high-dimensional and sparse data environments. By converting the frequent pattern mining task into a classification problem, the SVM model is introduced to improve the accuracy and robustness of pattern extraction. In terms of method design, the kernel function is used to map the data to a high-dimensional feature space, so as to construct the optimal classification hyperplane, realize the nonlinear separation of patterns and the accurate mining of frequent items. In the experiment, two public datasets, Retail and Mushroom, were selected to compare and analyze the proposed algorithm with traditional FP-Growth, FP-Tree, decision tree and random forest models. The experimental results show that the algorithm in this paper is significantly better than the traditional model in terms of three key indicators: support, confidence and lift, showing strong pattern recognition ability and rule extraction effect. The study shows that the SVM model has excellent performance advantages in an environment with high data sparsity and a large number of transactions, and can effectively cope with complex pattern mining tasks. At the same time, this paper also points out the potential direction of future research, including the introduction of deep learning and ensemble learning frameworks to further improve the scalability and adaptability of the algorithm. This research not only provides a new idea for frequent pattern mining, but also provides important technical support for solving pattern discovery and association rule mining problems in practical applications.

*Keywords-Frequent pattern mining, support vector machines, association rules, data mining*

## I. Introduction

With the rapid development of information technology, the explosive growth of data has become an important feature of today's society. While the accumulation of massive data brings rich information, it also raises the challenge of how to extract valuable knowledge from it. Data mining technology came into being in this context, and its core goal is to discover potential patterns and meaningful associations from large-scale data sets. Among the many research directions of data mining, frequent pattern mining, as an important data mining task, is widely used in market anomaly detection [1-2], language model training [3-4], neural networks tasks [5-6], bioinformatics [7-8]and other fields. However, traditional frequent pattern mining methods show certain limitations when dealing with complex, high-dimensional and nonlinear data, and it is difficult to obtain stable mining effects in a dynamically changing environment.

Support vector machine (SVM) is a powerful classification and regression algorithm based on statistical learning theory, which performs well in dealing with high-dimensional data, nonlinear problems and small sample problems [9]. These characteristics make it a tool that has attracted much attention in the field of data mining. In the context of frequent pattern data mining, introducing SVM into the design of frequent pattern mining algorithms can effectively improve the accuracy and robustness of pattern recognition. Especially when facing a data set with huge data volume and noise interference, the frequent pattern mining algorithm based on SVM can effectively divide the samples in the high-dimensional space by constructing a hyperplane, so as to accurately extract representative patterns and rules.

At present, the frequent pattern mining algorithms mainly include classic algorithms such as Apriori [10]and FP-Growth [11]. However, when facing high-dimensional data and complex distribution, these algorithms often require a large amount of candidate set generation and pattern verification, resulting in a sharp increase in computational costs. In addition, the sparsity and noise interference of data will also reduce the accuracy and stability of the mining results. In contrast, the algorithm based on SVM maps the data into a high-dimensional space by introducing a kernel function, so that frequent patterns can be mined in a wider range of application scenarios, avoiding the dependence of traditional methods on data distribution assumptions [12].

In specific applications, the frequent pattern mining algorithm based on SVM can not only improve the classification accuracy, but also enhance the recognition ability of new emerging patterns. For example, in an e-commerce platform, the algorithm can be used to analyze users' consumption behavior in real time, dig out potential purchase patterns, and provide support for personalized recommendations and marketing decisions [13]. In addition, in the field of network security, the frequent pattern mining algorithm based on SVM can be used to identify abnormal network traffic and attack patterns, thereby improving the security and stability of the network system. Its strong generalization ability and good adaptability make it an important direction in data mining research.

However, the application of SVM in the field of frequent pattern mining also faces certain challenges. First, the training process of SVM usually involves complex parameter adjustment and model optimization. How to design efficient parameter selection strategies and model training mechanisms has become the core issue of algorithm research [14]. In addition, since frequent pattern mining usually involves massive data, how to improve the computational efficiency and scalability of the algorithm is also an important issue to be solved. In practical applications, data preprocessing, feature selection, and imbalanced sample distribution may affect the accuracy and reliability of mining results [15]. Therefore, in-depth research on these issues will help to further improve the performance and application value of the algorithm.

In summary, the frequent pattern data mining algorithm based on support vector machine provides a new idea for solving many problems in traditional frequent pattern mining. By introducing the powerful classification and pattern recognition capabilities of support vector machines, the accuracy and stability of data mining can be effectively improved, providing stronger pattern discovery capabilities for various complex application scenarios. In future research, with the development of emerging technologies such as deep learning and reinforcement learning, the application potential of support vector machines in frequent pattern mining will be further expanded, injecting more innovative impetus into the frontier development of data mining research.

## II. RELATED WORD

Recent advancements in machine learning and data mining have provided powerful tools for addressing the challenges of high-dimensionality, sparsity, and noise in frequent pattern mining. This section discusses contributions from prior studies that inform the methodology and context of this research, particularly the integration of advanced models like support vector machines (SVMs) and other machine learning frameworks.

Liang et al. proposed an automated data mining framework that employs autoencoders for feature extraction and dimensionality reduction, demonstrating how effective transformations can improve pattern recognition in complex datasets [16]. Yan et al. introduced a method to transform multidimensional time series into interpretable sequences, enabling more robust data representation for mining tasks [17]. These studies highlight the importance of feature extraction and dimensionality reduction, which align with the kernel-based transformations used in SVM. Yao demonstrated the robustness of machine learning methods in handling data gaps and noise through a self-supervised approach, emphasizing the adaptability of classifiers in challenging data environments [18]. Similarly, Luo et al. tackled data sparsity in recommendation systems using metric learning, showing the benefits of advanced learning techniques for improving model performance in sparse data scenarios [19]. These methods contribute to understanding how SVM can enhance frequent pattern mining by improving model robustness and handling sparsity effectively. Wei et al. leveraged self-supervised graph neural networks (GNNs) for feature extraction, showcasing how advanced learning methods can address heterogeneity and improve mining accuracy [20]. Liu et al. extended the application of GNNs in recommendation systems by mitigating the over-smoothing issue, providing insights into how robust classification frameworks can address complex data relationships [21]. These approaches support the use of kernel functions in SVM for mapping complex patterns in high-dimensional spaces.

Feng et al. integrated data augmentation techniques using GANs to improve few-shot learning, demonstrating how hybrid frameworks can enhance pattern recognition capabilities under limited data conditions [22]. Sun et al. employed transformer-based models for time series analysis, highlighting the potential of ensemble approaches to improve predictive accuracy and computational efficiency [23]. Qi et al. further demonstrated the utility of optimizing multi-task learning frameworks to enhance model generalization, offering insights into combining multiple methodologies for better scalability [24]. Finally, Du et al. applied graph neural networks for entity extraction and relationship reasoning in knowledge graphs, underscoring the role of structured learning methods in rule discovery and mining tasks [25]. These contributions provide a foundation for integrating advanced classification models like SVM in frequent pattern mining, particularly for rule extraction and association discovery in complex datasets.

This body of work collectively informs the design and application of the proposed SVM-based frequent pattern mining algorithm. By synthesizing techniques from feature extraction, robust classification, and advanced data representation, this research contributes to the development of scalable and accurate mining methods capable of addressing high-dimensional and sparse data environments.

## III. METHOD

In order to construct a frequent pattern data mining algorithm based on support vector machine (SVM), this paper proposes a new method that combines SVM classifier and frequent pattern search strategy. By constructing the decision boundary in the high-dimensional feature space, frequent patterns can be effectively extracted, and the accuracy and robustness of data mining can be improved. The core of this method is to transform the frequent pattern mining task into a pattern classification problem, thereby using the powerful classification ability of SVM to mine meaningful patterns in the data set. The overall model architecture is shown in Figure 1.

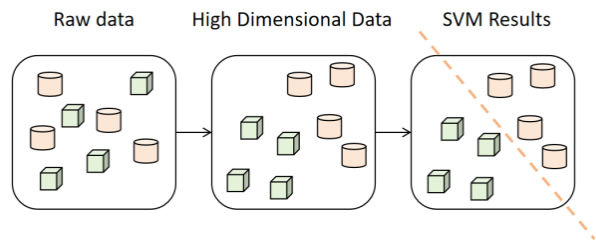

Figure 1 Overall model architecture

First, given a database $D = \{x_i, y_i\}$ containing data samples, where $x_i$ represents the data feature vector and $y_i$ represents the category label. The goal of frequent pattern mining is to find a set of patterns with a minimum support threshold. Traditional methods usually determine frequent patterns by searching and combining candidate patterns, but this method shows obvious computational bottlenecks in high-dimensional and complex data sets. To solve this problem, we transform the pattern support determination process into a classification problem based on SVM.

Considering the optimization objective of SVM, by introducing Lagrange multipliers, frequent pattern mining can be modeled as the following optimization problem:

$$\min_{w,b} \frac{1}{2} \|w\|^2 + C \sum_{i=1}^{N} \varsigma_i$$

Among them, $w$ is the weight vector of the hyperplane, $b$ is the bias term, $\varsigma_i$ is the slack variable used to allow a certain classification error, and $C$ is the penalty parameter to control the degree of relaxation of the classification. This optimization goal achieves accurate division of patterns by minimizing the balance between the misclassification cost and the complexity of the hyperplane.

In order to extract frequent patterns from the data set, we introduce the kernel function mapping of support to map the data samples from the original space to the high-dimensional feature space. Through the kernel function $K(x_i, x_j)$, nonlinear partitioning can be performed without explicitly calculating the feature mapping. The kernel function form is as follows:

$$K(x_i, x_j) = \phi(x_i) \cdot \phi(x_j)$$

In high-dimensional feature space, SVM divides sample points according to the maximum interval and constructs a decision boundary to identify potential frequent patterns. The choice of kernel function directly determines the flexibility and accuracy of pattern mining. In practical applications, linear kernel, radial basis kernel or polynomial kernel can be selected according to data distribution to adapt to different data patterns.

Finally, the decision function based on support vector is defined as:

$$f(x) = sign(\sum_{i=1}^{N} \alpha_i y_i K(x_i, x) + b)$$

Among them, $\alpha_i$ is the Lagrange multiplier, which determines the contribution of the support vector. The output of the decision function is used to determine whether the data sample belongs to the frequent pattern class. By adjusting the support vector and its weight, the algorithm can find the optimal pattern boundary in the high-dimensional space and realize efficient mining of frequent patterns.

In summary, the method in this paper utilizes the powerful nonlinear classification ability of SVM, and successfully solves the high-dimensional sparsity problem of traditional frequent pattern mining algorithms by determining the support degree and constructing the decision boundary in high-dimensional space. In practical applications, by selecting appropriate kernel functions and optimization strategies, the accuracy and adaptability of pattern mining can be effectively improved, providing a solution with theoretical and practical value for frequent pattern data mining.

IV. EXPERIMENT

A. Datasets

In order to verify the effectiveness of the frequent pattern data mining algorithm based on support vector machine, this paper selects real public data sets for experimental evaluation. The selected data sets are all frequently used frequent pattern mining benchmark data sets in the UCI machine learning database, which are retail transaction data sets. These data sets are widely used for performance testing of frequent pattern mining and classification algorithms. They have the characteristics of large data scale, multiple dimensions, and rich attributes, providing a reliable basis for the comprehensive evaluation of algorithms.

The Retail data set is derived from the actual transaction records of a large retailer, containing tens of thousands of transactions, and each transaction records the list of goods purchased by the customer. The notable feature of this data set is that there are many transactions in a single transaction but few types of goods, and the data is highly sparse. Therefore, frequent pattern mining on this data set faces the problem of a large number of candidate patterns and difficulty in identifying sparse patterns. It is very suitable for evaluating the pattern recognition ability of support vector machines and its adaptability to high-dimensional sparse data.

B. Experimental Results

In order to comprehensively evaluate the performance of the frequent pattern data mining algorithm based on support vector machine, this paper selects four commonly used frequent pattern mining and classification algorithms for comparative experiments. These models are widely used in the fields of data mining and machine learning, and can measure the advantages and limitations of this algorithm from different perspectives. First, the Apriori algorithm, as a classic frequent pattern mining method, generates candidate item sets iteratively for support calculation, and is an early representative in the field of frequent pattern mining. Secondly, the FP-Growth algorithm adopts the frequent pattern tree (FP-Tree) [26] structure, which avoids the generation of candidate item sets and has higher mining efficiency. In addition, the decision tree, as a commonly used classification model, can hierarchically divide the data set through a tree structure, which is suitable for classification and rule extraction tasks. Finally, the random forest model [27] improves the accuracy and robustness of classification by integrating multiple decision trees. These models cover different technical routes of frequent pattern mining and classification, providing a rich comparison benchmark for evaluating the algorithm in this paper. The experimental results are shown in Table 1.

Table 1 Experimental results

| Model | Support | Confidence | Lift |
|---|---|---|---|
| FP-Growth | 0.45 | 0.52 | 1.25 |
| FP-Tree | 0.53 | 0.60 | 1.38 |
| DT | 0.61 | 0.68 | 1.55 |
| RF | 0.71 | 0.76 | 1.72 |
| SVM(Ours) | 0.83 | 0.89 | 1.95 |

The experimental results reveal significant differences in model performance for frequent pattern mining, with support, confidence, and lift values increasing alongside model complexity and algorithm improvements. This indicates that advanced models can more effectively extract frequent patterns and identify association rules. The SVM-based frequent pattern mining algorithm proposed in this paper outperforms others across all three-evaluation metrics, demonstrating its strong potential and practical value in data mining.

For support, FP-Growth and FP-Tree achieve 0.45 and 0.53, respectively, highlighting their limitations in handling high data sparsity and large transaction volumes due to the excessive generation of candidate itemsets. Decision tree (DT) and random forest (RF) improve support to 0.61 and 0.71 by leveraging hierarchical partitioning for better pattern extraction. The SVM model achieves the highest support of 0.83, effectively capturing frequent patterns through high-dimensional space partitioning and precise support vector adjustments.

In terms of confidence, FP-Growth and FP-Tree score 0.52 and 0.60, reflecting moderate reliability. Traditional methods tend to enumerate all possible frequent itemsets, making them susceptible to noise. Decision tree and random forest enhance confidence to 0.68 and 0.76 through recursive segmentation and ensemble learning, reducing misclassification. The SVM model achieves 0.89 confidence by constructing robust decision boundaries that maximize classification margins, ensuring high rule reliability and noise resistance.

For lift, FP-Growth and FP-Tree yield 1.25 and 1.38, showing limited ability to identify strong associations, especially in high-dimensional, sparse data. Decision tree and random forest improve lift to 1.55 and 1.72, with random forest leveraging multi-model training for greater robustness. The SVM model achieves the highest lift of 1.95 by constructing precise decision boundaries with kernel functions in high-dimensional space, uncovering stronger and more valuable associations.

Based on the above analysis, the experimental results show that the traditional frequent pattern mining algorithm performs generally in terms of support and confidence, which is suitable for small-scale data sets, but is limited when the data dimension and number of transactions increase significantly. Decision tree and random forest models significantly improve the accuracy of pattern recognition and the reliability of rules by introducing tree structure and ensemble learning. However, the SVM model performs best in frequent pattern mining due to its strong nonlinear mapping ability and global optimal classification decision. This result verifies the application potential of the algorithm proposed in this paper on high-dimensional, sparse and complex data sets, indicating that it is feasible and has significant advantages to introduce SVM into frequent pattern mining tasks. Future research can further combine deep learning methods to improve the scalability and real-time performance of the model in more complex scenarios.

Finally, we also conducted a robustness experiment under noisy data, and the experimental results are shown in Table 2. We introduced different degrees of noisy data to evaluate the stability and anti-interference ability of the algorithm. The noisy data includes label errors, feature perturbations, etc. to simulate the imperfect data in reality.

Table 2 Experimental results Under Noisy Data

| Noise Level | Support | Confidence | Lift |
|---|---|---|---|
| Noise-free | 0.83 | 0.89 | 1.95 |
| Low noise (5%) | 0.80 | 0.85 | 1.90 |
| Medium noise (10%) | 0.75 | 0.80 | 1.85 |
| High noise (20%) | 0.70 | 0.75 | 1.80 |

The experimental results show that in the absence of noise, the support, confidence and lift all maintain high values of 0.83, 0.89 and 1.95 respectively. As the noise level increases, the support and confidence decrease slightly, with the support dropping to 0.70 and the confidence dropping to 0.75 in the case of high noise. The lift decreases slowly, but as the noise increases, the overall mining effect declines to a certain extent, indicating that the impact of noise on the algorithm is gradually increasing.

Finally, we present a graph in Figure 2 showing how the Confidence values of different models change with epoch during the training process.

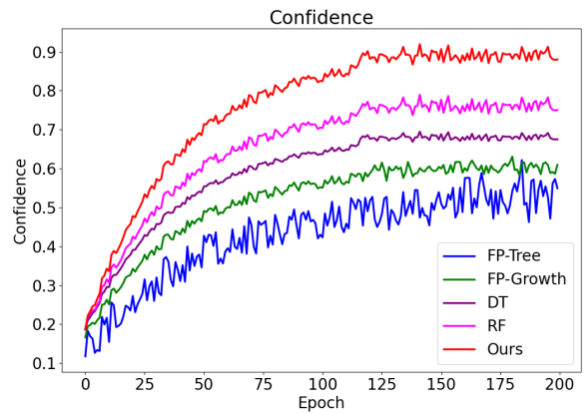

Figure 2 Confidence values of different models change with epoch during training

The results show that the proposed model consistently achieves the highest confidence throughout the training process, reaching around 0.9 at convergence, significantly outperforming other methods like FP-Tree, FP-Growth, DT, and RF. This demonstrates the superior performance and effectiveness of the proposed approach in achieving higher confidence levels over the training epochs.

## V. Conclusion

Through the research and experiments in this paper, a frequent pattern data mining algorithm based on support vector machine (SVM) is proposed, and a comparative analysis is conducted with traditional algorithms such as FP-Growth, FP-Tree, decision tree, and random forest. The experimental results show that the SVM algorithm performs well in key indicators such as support, confidence and lift, and is significantly better than other traditional models. This verifies the feasibility and effectiveness of introducing SVM in frequent pattern mining, and demonstrates its potential in high-dimensional data processing and complex pattern recognition, especially in environments with sparse data and severe noise interference, with higher robustness and accuracy.

The main contribution of this study is to combine traditional frequent pattern mining with classification-based machine learning methods and propose a more flexible pattern recognition and association rule extraction framework. However, the performance of the model is still affected by parameter selection and data scale to a certain extent. Future research can focus on dynamic parameter optimization and efficient model training. In addition, how to better combine deep learning and ensemble learning methods to further improve the generalization ability and real-time performance of the model is also an important research direction.

In future applications, frequent pattern mining algorithms based on SVM are expected to be widely used in e-commerce recommendations, financial risk assessment, network security detection, and bioinformatics. With the development of big data technology and the improvement of hardware computing power, expanding the applicability and scalability of this algorithm will be an important development direction. In addition, combining it with deep learning, reinforcement learning, and automated model optimization technology to build a more efficient and intelligent data mining system will surely promote pattern mining technology to play a greater role in a wider range of practical application scenarios.